\pdfoutput=1
\documentclass[11pt]{article}

\usepackage[review]{acl}

\usepackage{times}
\usepackage{latexsym}

\usepackage[T1]{fontenc}
\usepackage[utf8]{inputenc}
\usepackage{graphicx}

\usepackage{microtype}

\usepackage{inconsolata}

\title{Song Emotion Classification of Lyrics with Out-of-Domain Data under Label Scarcity}

\author{Jonathan Sakunkoo \\
  Stanford University OHS \\
  415 Broadway, Redwood City \\
  CA 94063, USA \\
  jonkoo@ohs.stanford.edu \\\And
  Annabella Sakunkoo \\
  Stanford University OHS \\
  415 Broadway, Redwood City \\
  CA 94063, USA \\
  apianist@ohs.stanford.edu \\
   }

\begin{document}

\maketitle

\section{Introduction}
 
Songs have been found to profoundly impact human emotions, with lyrics having significant power to stimulate emotional changes in the audience (Stratton and Zlanowski, 1994; Greer et al., 2019; Linnemann et al., 2015; Anderson et al., 2003). While lyrics emotion classification has immense potential benefits, major music platforms still lack the ability to classify song emotions based on lyrics. Furthermore, there is a scarcity of large, high quality in-domain datasets for lyrics-based song emotion classification (Edmonds and Sedoc, 2021; Zhou, 2022). It has been noted that in-domain training datasets are often difficult to acquire (Zhang and Miao, 2023) and that label acquisition is often limited by cost, time, and other factors (Azad et al., 2018). This abstract explores a CNN system to classify song emotions based on a large out-of-domain dataset as a creative, novel solution to the challenge of data scarcity in the domain of song lyrics. We find that models trained on a large Reddit comments dataset achieve satisfactory performance and are generalizable to lyrical emotions, thus giving insights into and a promising possibility in leveraging large, publicly available out-of-domain datasets for domains whose in-domain data are lacking or costly to acquire. 

\section{Related Work}
Many researchers have attempted to classify song emotions. Nuzzolo (2015) classified music mood with harmony, rhythm, and spectral features. Similarly, Tong (2022) classified song emotions based on pitch frequency and band energy distribution, not lyrics. Some researchers have used lyrics to classify emotions such as classifying Chinese songs (Jia 2022). Lee and Yang (2009), Mihalcea and Strapparava (2012), and Choi et al. (2018) attempted to classify song emotions but did not use neural networks. Studies by Mi-
halcea and Strapparava (2012) and Edmonds and Sedoc (2022) labeled song emotions based on Ekman’s 6 core emotions of anger, disgust, fear, joy, sadness, and surprise (Ekman 1993) and Plutchik’s 8 core emotions of anger, anticipation, disgust, fear,
joy, sadness, surprise, and trust (Plutchik, 2001), respectively. Edmonds and Sedoc (2022) found that models trained on large corpora of tweets and 2-
person conversations do not quite generalize to lyrical data for emotions and found all models trained on both in-domain and out-of-domain data to have low accuracy when classifying certain emotions such as "surprise" and "disgust."

\section{Approach}
\subsection{Dataset}
With the lack of large, high-quality datasets that focus on song-emotion classification, using out-of-domain data can have several potential benefits. For example, leveraging publicly available out-of-domain data can mitigate data scarcity issues when obtaining a large amount of labeled data in the target domain is challenging or expensive. Training on out-of-domain, diverse data can also improve the model's ability to generalize to new, unseen data and be more robust. There exist certain large, publicly available datasets containing text strings pre-categorized into specific emotional classifications. For example, the Kaggle Dataset, available on the Kaggle platform, classifies almost 40,000 tweets into 13 distinct emotions (Gupta, 2020). The GoEmotions dataset classifies over 58,000 Reddit user comments into 27 emotional categories (Demszky et al., 2020). EmotionLines is a dataset that labels 29,245 utterances from TV series "Friends" and Facebook messages into 6 emotions (Chen et al., 2018).

The Reddit comments in GoEmotions are diverse and potentially generalizable to lyrics in several ways. Unlike tweets or Facebook messages (Chen et al., 2018), which are very short, Reddit comments are more comparable to lyrics in length. A Reddit comment also presents a one-person perspective, similar to lyrics, as opposed to two-person dialog studied by Edmonds and Sedoc (2022). Although they may differ in content and purpose, using Reddit comments also has a potential advantage as training data for our purpose as it is likely that humans are not born to communicate particularly in lyrical language, and thus their interpretations and perceptions of lyrics are based on the context and familiarity of conversational language. Thus, while the Reddit dataset does not include music lyrics, we hypothesize that it might be large and diverse enough to train deep models to accurately predict emotions in song lyrics. 
The basic emotion categories in our study are: "anger", "confusion", "desire", "fear", "grief", "excitement", "love", and “sadness" (Since surprise and disgust have been found to have low accuracy (Edmonds and Sedoc, 2022), we excluded these emotions.) 

\subsection{Implementation}

We implemented a sequential neural network model explicitly tailored for emotion classification tasks, using text. This architecture strove to discern the underlying emotional context within textual sequences by leveraging both embedding and convolutional methods. The structure was as follows:

Flattening Layer: Given the sequential nature of text data, the model initiated by flattening the input sequences. The transformation ensured compatibility with the subsequent embedding layer.

Word Embedding: To transform raw text into meaningful vector representations, an embedding layer mapped each word to a dense vector within a space defined by embed\_size. This layer captured semantic relationships between words, and unlike scenarios where embeddings might be pre-trained (like GloVe or Word2Vec), here, they were learned alongside the model, honing them for emotion-based distinctions.

Dropout for Regularization: To combat overfitting, especially given the intricacies of emotional nuances in text, a dropout layer was introduced with a 20\% drop rate. This mechanism periodically nullified certain input units, promoting robustness.

Convolutional Feature Extraction: The architecture employed two 1D convolutional layers, each configured with 100 filters of kernel size 4. By sliding these filters over the embedded textual sequences, the model grasped local patterns and potential emotional markers present in word combinations.

Pooling Layers: After feature extraction, a max-pooling layer condensed the spatial dimensions by retaining only the predominant emotional markers. Further abstraction was achieved through a global max-pooling layer, which summarized the entirety of the sequence into its most pronounced features.

Classification through Dense Layers: The high-level emotional features then traversed a series of fully connected layers. A dense layer with 64 units, followed by another with 32 units—both adopting the "softplus" activation—processed these features, honing in on the underlying emotions. An intermediary dropout layer, with a 20\% rate, acted as a safeguard against overfitting.

Output Layer for Emotion Labels: The result was an output dense layer with units equivalent to the 8 emotions. With a "sigmoid" activation, the model seemed adept at multi-label emotion classification, enabling it to detect multiple emotions that might coexist within a given lyric.

Finally, we employed the binary cross-entropy loss function and ADAM optimizer.  Overall, our model harnessed the power of convolutional layers to grasp the local context within sequences. However, unlike architectures that might employ recurrent layers or multi-channel embeddings, this design emphasized simplicity and direct feature extraction.

After training the model on 58,000 Reddit comments, we ran the model on 100 songs from Billboard's Popular Chart. Twenty participants, all native speakers of English, labeled each song with 3 emotions from their perspective. The labels were then compared with the model's results, and the overlaps of the human-assigned labels and the system's classified emotions were measured. 

\section{Results and Conclusion}
The accuracy as measured by the percentage of overlaps of human labels and system classifications is 88\%. Samples of the classification results are in Appendix A. Previous work has found Tweet and dialog data not quite generalizable for lyrical data. Our study presents the novelty of using a different out-of-domain dataset and finds that CNN models trained on Reddit comments may yield satisfactory accuracy for song lyrics emotion classification. This work contributes to lyrical emotion classification and computational linguistics as it provides insights into the possibility of using large, publicly available out-of-domain data as an alternative to scarce, hard-to-acquired in-domain datasets.

\nocite{Abrams, Anderson, Chavez, Demsky, Greer, Jia, Lerner, Linnemann, Nuzzolo, Rosenberg, Chen, Tong, Choi, Mihalcea, Lee, Edmonds, Ekman}

\bibliography{custom}

\appendix\textbf{Appendix A}

Figure 1 below shows samples of song emotion classification. The model classified “Sweet Dreams” by Alan Walker as a combination of love and excitement at 30\% and 19\% respectively. For “i miss u,” the model classified the song as a mixture of love and sadness, corresponding to the participants' labels. “They Don’t Really Care about Us” was classified as mostly sadness, followed by anger and desire. It classified “California Girls” by The Beach Boys as mostly desire. The model's performance was re-evaluated by human participants to add a layer of practical validation. Qualitatively, the participants felt that the model’s classification generally made sense. For example, “They Don’t Really Care about Us” conveyed a sense of dissatisfaction, anger, and disappointment about injustice. The participants also felt that the classification of “California Girls” as mostly desire was right and interesting.
\begin{figure}[htp]
    \centering
    \includegraphics[width=5cm]{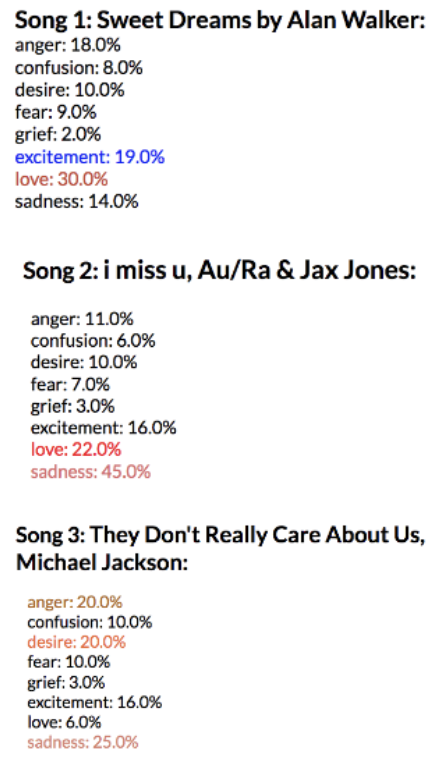}
    \centering
\includegraphics[width=5cm]{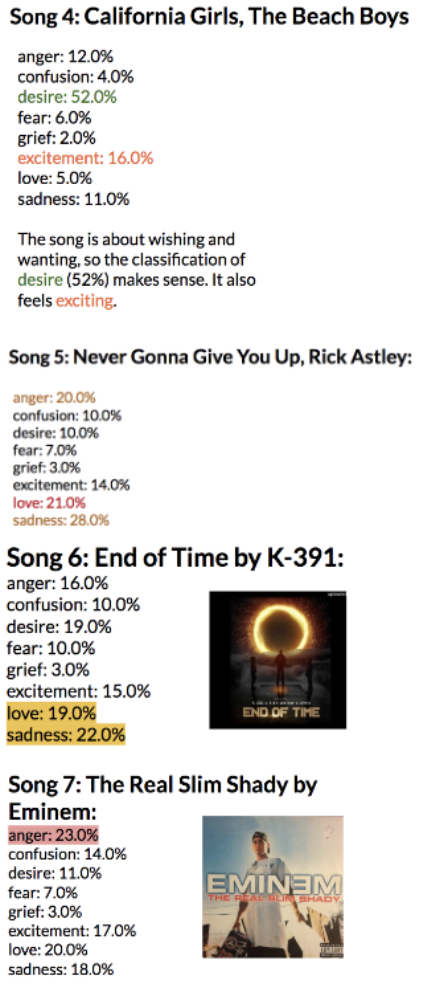}
    \caption{Sample songs with classification results}
\end{figure}

\end{document}